\documentclass[10pt,twocolumn,letterpaper]{article}
\usepackage{iccv}
\usepackage{times}
\usepackage{epsfig}
\usepackage{graphicx}
\usepackage{amsmath}
\usepackage{amssymb}
\usepackage{subfig}
\usepackage{rotating}
\usepackage{balance}
\usepackage{multirow}

\usepackage[pagebackref=true,breaklinks=true,letterpaper=true,colorlinks,bookmarks=false]{hyperref}

 \iccvfinalcopy 

\def\iccvPaperID{2490} 

\ificcvfinal\fi

\newcommand{\hide}[1]{}

\newcommand{\dquote}[1]{``#1''}

\begin{document}

\title{Deep Generative Adversarial  Compression Artifact Removal}

\author{Leonardo Galteri, Lorenzo Seidenari, Marco Bertini, Alberto Del Bimbo\\
MICC, University of Florence\\
{\tt\small https://www.micc.unifi.it/}
}
\maketitle

\begin{abstract}
Compression artifacts arise in images whenever a lossy compression algorithm is applied. These artifacts eliminate details present in the original image, or add noise and small structures; because of these effects they make images less pleasant for the human eye, and may also lead to decreased performance of computer vision algorithms such as object detectors. To eliminate such artifacts, when decompressing an image, it is required to recover the original image from a disturbed version. 
To this end, we present a feed-forward fully convolutional residual network model trained using a generative adversarial framework. To provide a baseline, we show that our model can be also trained optimizing the Structural Similarity (SSIM), which is a better loss with respect to the simpler Mean Squared Error (MSE).

 Our GAN is able to produce  images with more photorealistic details than MSE or SSIM based networks. Moreover we show that our approach can be used as a pre-processing step for object detection in case images are degraded by compression to a point that state-of-the art detectors fail. In this task, our  GAN method obtains better performance than MSE or SSIM trained networks.
\end{abstract}

\section{Introduction}
\begin{figure}
\centering
\includegraphics[width=\columnwidth]{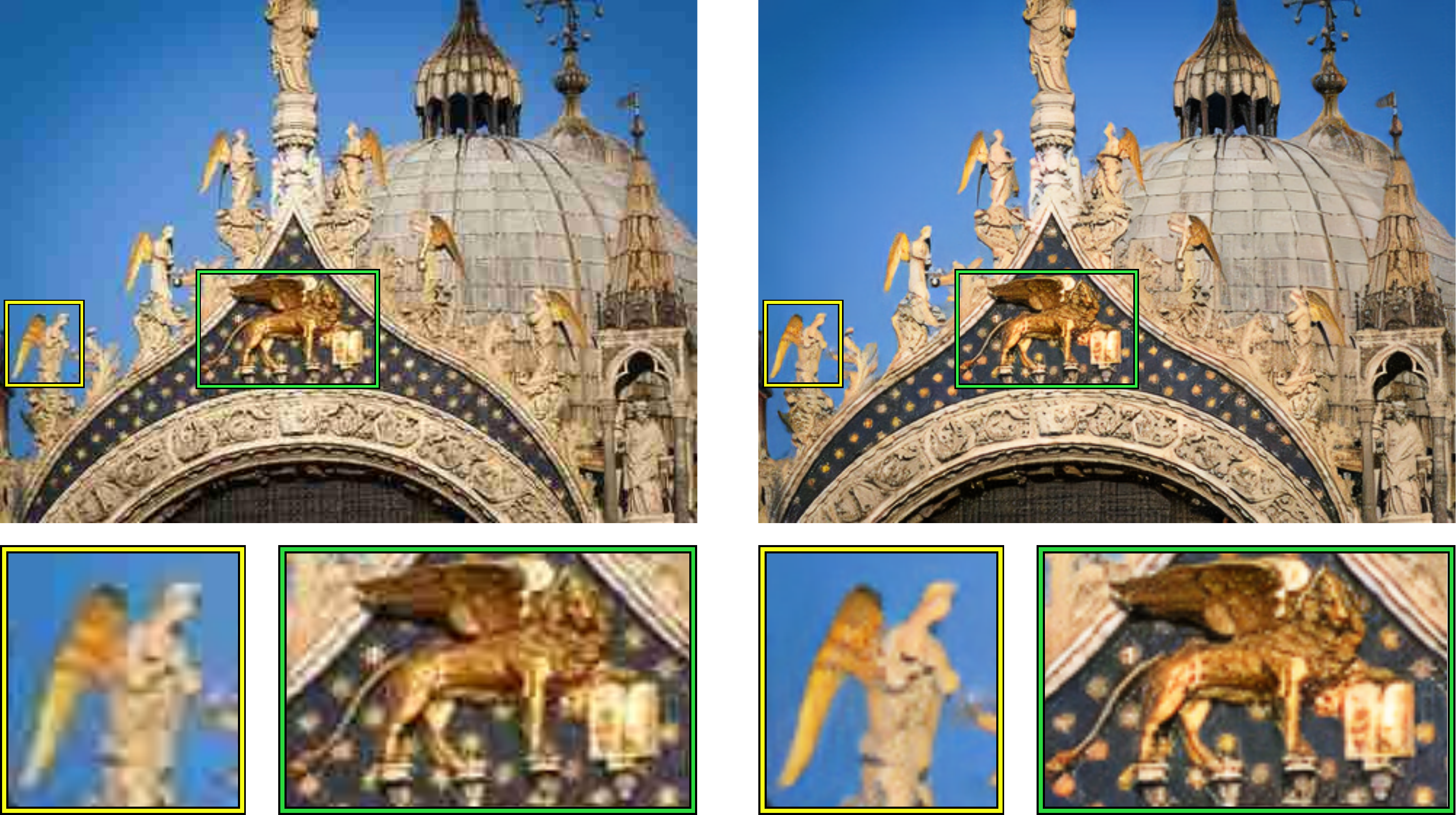}
\caption{Left: A JPEG compressed image with two highlights of degraded regions. Right: our reconstruction where both regions are consistently sharper and most artifacts are removed. Best viewed in color on computer screen.\label{fig:eyecatcher}}\vspace{-24pt}
\end{figure}

Image and video compression algorithms are commonly used to reduce the dimension of media files, to lower their storage requirements and transmission time. These algorithms often introduce compression artifacts, such as blocking, posterizing, contouring, blurring and ringing effects, as shown in Fig.~\ref{fig:eyecatcher}. Typically, the larger the compression factor, the stronger is the image degradation due to these artifacts. However, simply using images with low compression is not always a feasible solution: images used in web pages need to be compressed as much as possible to speed up page load to improve user experience. 
Many types of video/image streams are necessarily requiring low-bandwidth, e.g.~for drones, surveillance and wireless sensor networks. Also in tasks such as entertainment video streaming, like Netflix, there is need to reduce as much as possible the required bandwidth, to avoid network congestions and to reduce costs.
So far, the problem of compression artifact removal has been treated using many different techniques, from optimizing DCT coefficients \cite{Zhang-2013} to adding additional knowledge about images or patch models \cite{Liu-2015}; however the very vast majority of the many works addressing the problem have not considered convolutional neural networks (CNN). To the best of our knowledge CNNs have been used recently to address artifact reduction only in two works \cite{dong2015compression, svoboda2016compression}, while another work has addressed just image denoising \cite{Zhang-2017}.
These techniques have been successfully applied to a different problem of image reconstruction,  that is super-resolution, to reconstruct images from low resolution, adding missing details to down-sampled images \cite{Ledig-2016}.

In this work we address the problem of artifact removal using convolutional neural networks. The proposed approach can be used as a post-processing technique applied to decompressed images, and thus can be applied to different compression algorithms such as JPEG, intra-frame coding of H.264/AVC and H.265/HEVC. 

To evaluate the quality of reconstructed images, after artifact removal, there is need to evaluate both subjective and objective assessments. The former are important since most of the time a human will be the ultimate consumer of the compressed media. The latter are important since obtaining subjective evaluations is slow and costly, and the goal of objective metrics is to  predict perceived image and video quality automatically. Peak Signal-to-Noise Ratio (PSNR) and Mean Squared Error (MSE) are the most widely used objective image quality/distortion metrics. However, they have been criticized because they do not correlate well with perceived quality measurement \cite{Wang-2002}. To face these issues, Structural Similarity index (SSIM) has been proposed \cite{wang2004image}.
 Finally, we can expect that more and more viewers will be computer vision systems that automatically analyze media content, e.g.~to interprete it in order to perform other processing. To consider also this scenario we have to assess the performance of computer vision algorithms when processing reconstructed images.

In this work we show how deep CNNs can be used to remove compression artifacts by directly optimizing SSIM on reconstructed images, showing how this approach leads to state-of-the-art result on several benchmarks. However, although SSIM is a better model for image quality than PSNR or MSE, it is still too simplistic and insufficient to capture the complexity of the human perceptual system.
Therefore, to learn better reconstructive models, we rely on a Generative Adversarial Network where there is no need to specify a loss directly modeling image quality.

We have performed different types of experiments, to assess the diverse benefits of the different types of networks proposed in this paper, using subjective and objective assessments.  Firstly, we show that not only SSIM objective metric is improved, but also that performance of object detectors improve on highly compressed images; this is especially true for GAN artifact removal. Secondly, according to human viewers our GAN reconstruction has a higher fidelity to the uncompressed versions of images.

\section{Related Work}
Removing compression artifacts has been addressed in the past. There is a vast literature of image restoration, targeting image compression artifacts. The vast majority of the approaches can be classified as processing based~\cite{foi2007pointwise,wong2009document,yang2000blocking,Zhang-2013,Li-2014, Chang-2014, Zhang-2016, Dar-2016} and a few ones are learning based \cite{dong2015compression,svoboda2016compression,MaoSY16a,wang2016d3}.

Processing based methods typically rely on information in the DCT domain.  Foi \etal\cite{foi2007pointwise} developed SA-DCT, proposing to use clipped or attenuated DCT coefficients to reconstruct a local estimate of the image signal within an adaptive shape support. Yang \etal\cite{yang2000blocking}, apply a DCT-based lapped transform directly in the DCT domain, in order to remove the artifacts produced by quantization. Zhang \etal\cite{Zhang-2013}, fuse two predictions to estimate DCT coefficients of each block: one prediction is based on quantized values of coefficients and the other is computed from nonlocal blocks coefficients as a  weighted average.  Li \etal\cite{Li-2014} eliminate artifacts due to contrast enhancement, decomposing images in structure and texture components, then eliminating the artifacts that are part of the texture component. Chang \etal\cite{Chang-2014} propose to find a sparse representation over a learned dictionary from a training images set, and use it to remove the block artifacts of JPEG compression images. Dar \etal\cite{Dar-2016} propose to reduce artifacts by a regularized restoration of the original signal. The procedure is formulated  as a regularized inverse-problem for estimating the original signal given its reconstructed form, and  the nonlinear compression-decompression process is approximated by a linear operator, to obtain a tractable formulation. The main drawback of these methods is that they usually over-smooth the reconstructed image. Indeed it is hardly possible to add consistent details at higher frequencies with no semantic cues of the underlying image.

Learning based methods have been proposed following the success of deep convolutional neural networks (DCNN). The basic idea behind applying a DCNN to this task is to learn an image transformation function that given an input image will output a restored version. Training is performed by generating degraded versions of images which are used as samples for which the ground truth or target is the original image. The main advantage of learning based methods is that, since they are fed with a large amount of data they may estimate accurately an image manifold, allowing an approximated inversion of the compression function. This manifold is also aware of image semantics and does not rely solely on DCT coefficient values or other statistical image properties.
Dong \etal\cite{dong2015compression} propose artifact reduction CNN (AR-CNN) which is based on their super-resolution CNN (SRCNN); both models share a common structure, a feature extraction layer, a feature enhancement layer, a non-linear mapping and a reconstruction layer. The structure is designed following sparse coding pipelines. Svoboda \etal\cite{svoboda2016compression} report improved results by learning a feed-forward CNN to perform image restoration; differently from \cite{dong2015compression} the CNN layers have no specific functions but they combine residual learning, skip architecture and symmetric weight initialization to get a better reconstruction quality.
 
 Similar approaches have been devised, to target different image transformation problems, such as image super-resolution~\cite{BrunaSL15,Ledig-2016, johnson2016perceptual, Dahl-2017}, style-transfer \cite{gatys2016image,johnson2016perceptual} and image de-noising \cite{Zhang-2017}.
 Zhang~\etal\cite{Zhang-2017} have recently addressed the problem of image denoising, proposing a denoising convolutional neural networks (DnCNN) to eliminate Gaussian noise with unknown noise level and showing  that residual learning (used in a single residual unit of the network) and batch normalization are beneficial for this task. The proposed network obtains promising results also on other denoising tasks such as super resolution and JPEG deblocking. 
 Gatys \etal\cite{gatys2016image} have shown that optimizing a loss accounting for style similarity and content similarity it is possible to keep the semantic content of an image and alter its style, which is transferred from another source. Johnson \etal\cite{johnson2016perceptual} propose a generative approach to solve style transfer, building on the approach of Gatys~\etal. Their method improves in terms of performance with respect of \cite{gatys2016image}, since the optimization is performed beforehand, for each style, it is possible to apply the transformation in real-time. Interestingly, with a slight variation on the learning, their method also can solve super-resolution. Kim \etal\cite{kim2016accurate} use a deeper architecture \cite{SimonyanZ14a} trained on residual images applying gradient clipping to speed-up learning. 
 Bruna \etal\cite{BrunaSL15} addressed super-resolution learning sufficient statistics for the high-frequency component using a CNN, Ledig \etal\cite{Ledig-2016} used a deep residual convolutional generator network, trained in an adversarial fashion. Dahl \etal\cite{Dahl-2017} propose to use a PixelCNN architecture for super-resolution task, applying it to magnification of $8\times8$ pixel images. Human evaluators have indicated that samples from this model look more photo realistic than a pixel-independent L2 regression baseline. 

We make the following contributions. We define a deep convolutional residual generative network~\cite{HeZRS15}, that we train with two strategies. Similarly to \cite{svoboda2016compression} our network is fully convolutional and is therefore able to restore images of any resolution. Differently from~\cite{svoboda2016compression} we avoid MSE loss and we use a loss based on SSIM, this improves results perceptually. Nonetheless, as also happening in the super-resolution task, networks trained to optimize the MSE produce overly smoothed images; this behavior unfortunately is also present  in our SSIM trained feed-forward network.

Generative adversarial networks \cite{goodfellow2014generative}, are instead capable of modeling complex multi-modal distributions and are therefore known to be able to generate sharper images. We propose an improved generator, trained in an adversarial framework. To the best of our knowledge we are the first proposing GANs to recover from compression artifacts. We use a conditional GAN \cite{mirza2014conditional}, to allow the generator to better model the artifact removal task.  An additional relevant novelty of this work is the idea of learning the discriminator over sub-patches of a single generated patch to reduce high frequency noise, such as mosquito noise, which instead arises when using a discriminator trained on the whole patch.

\section{Methodology}
In the compression artifact removal task the aim is to reconstruct an image $I^{RQ}$ from a compressed input image $I^{LQ}$. In this scenario, $I^{LQ} = A\left(I^{HQ}\right)$ is the output image of a compression algorithm $A$ with $I^{HQ}$ as uncompressed input image. Typically compression algorithms work in the YCrCb color space (e.g.~JPEG, H.264/AVC, H.265/HEVC), to separate luminance from chrominance information, and sub-sample chrominance, since the human visual system is less sensitive to its changes. For this reason, in the following, all images are converted to YCrCb and then processed. 

We describe $I^{RQ}$, $I^{LQ}$ and $I^{HQ}$ by real valued tensors with dimensions $W \times H \times C$, where C is the number of image channels. Certain quality metrics are evaluated using the luminance information only; in those cases all the images are transformed to gray-scale considering just the luminance channel $Y$ and $C=1$. Of course, when dealing with all the YCrCb channels $C=3$.

An uncompressed image $I^{HQ} \in [0,255]^{W \times H \times C}$ is compressed by:

\begin{equation}
I^{LQ} = A\left(I^{HQ},QF\right) \in [0,255]^{W \times H \times C}
\end{equation}
using a compression function $A$, using some quality factor $QF$ in the compression process. The task of compression artifacts removal is to provide an inverse function $G\approx A_{QF}^{-1}$ reconstructing $I^{HQ}$ from $I^{LQ}$:
\begin{equation}
G\left(I^{LQ}\right) = I^{RQ} \approx I^{HQ}
\end{equation}
where we do not include the QF parameter in the reconstruction algorithm since it is desirable that such function is independent from the compression function parameters.

To achieve this goal, we train a convolutional neural network $G\left(I^{LQ};\theta_g\right)$ with $\theta_g=\{W_{1:K};b_{1:K}\}$ the parameters representing weights and biases of the $K$ layers of the network. Given $N$ training images we optimize a custom loss function $ l_{AR} $ by solving:\vspace{-8pt}

\begin{equation}
\hat{\theta}_g = \arg \min_{\theta} \frac{1}{N}\sum_{n=1}^{N}l_{AR}\left( I^{HQ},G\left(I^{LQ},\theta_g\right)\right)
\end{equation} 

Removing compression artifacts can be seen as an image transformation problem, similarly to super-resolution and style-transfer. This category of tasks is conveniently addressed using generative approaches, i.e. learning a fully convolutional neural network (FCN)~\cite{long2015fully} able to output an improved version of some input. FCN architectures are extremely convenient in image processing since they perform local non-linear image transformations, and can be applied to images of any size. We exploit this property to speed-up the training process: since the artifacts we are interested in appear at small scales (close to the block size), we can learn from smaller patches, thus using larger batches.

We propose a generator architecture that can be trained with direct supervision or combined with a discriminator network to obtain a generative adversarial framework. In the following we detail the network architectures that we have used and the loss functions devised to optimize such networks in order to obtain high quality reconstructions.

\subsection{Generative Network}\label{sec:generative}
In this work we use a deep residual generative network, which contains just blocks of convolutional layers and LeakyReLU non-linearities. 

\begin{figure}[!t]
\centering
\textbf{Generator Network}\par\bigskip
\includegraphics[scale=0.65]{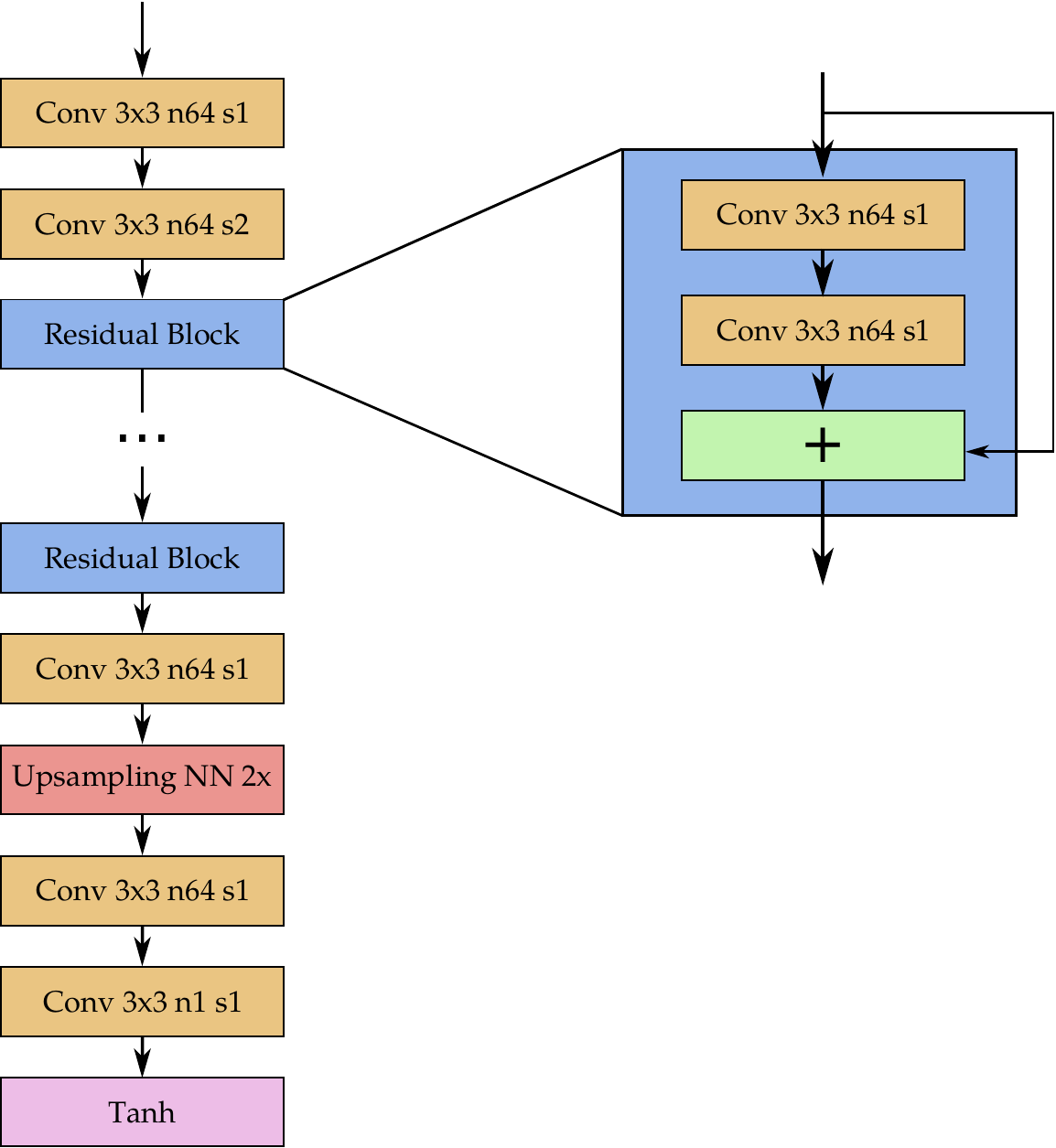}\par\bigskip
\caption{Architecture of Generator Network indicating with $n$ the number of filters and with $s$ the stride value for each Convolutional Layer.}\vspace{-12pt}
\label{fig:resnet}
\end{figure}

The architecture, shown in Fig.~\ref{fig:resnet}, is inspired by \cite{HeZRS15}.  Specifically, we use convolutional layers with $ 3 \times 3 $ kernels and 64 feature maps. Each convolutional layer is followed by a LeakyReLU activation. To reduce the overall number of parameters and to speed up the training time,  we first use a convolution with stride 2 to obtain the feature maps half the original size, and finally we employ a nearest-neighbor upsampling as suggested in \cite{odena2016deconvolution} to get the feature maps with original dimensions. We apply a padding of 1 pixel after every convolution, in order to keep the image size across the 15 residual blocks. We use replication as padding strategy in order to moderate border artifacts.

 We add another convolutional layer after the upsampling layer to minimize potential artifacts generated by the upsampling process. The last layer is a simple convolutional layer with one feature map followed by a \textit{tanh} activation function, in order to keep all the values of the reconstructed image in the range $ [-1,1]$ making the output image comparable with the input which is rescaled so to have values in the same range.

\subsection{Loss Functions for Direct Supervision}
In this section we deal with learning a generative network with a direct supervision, meaning that the loss is computed as a function of the reconstructed image $I^{RQ}$ and the target original image $I^{HQ}$. Weights are updated with a classical backpropagation.

\subsubsection{Pixel-wise MSE Loss}
As a baseline we use the Mean Squared Error loss (MSE):\vspace{-4pt}

\begin{equation}
l_{MSE} = \frac{1}{WH}\sum_{x=1}^{W}\sum_{y=1}^{H}\left(I^{HQ}_{x,y} - I_{x,y}^{RQ}\right)^2.
\end{equation}
This loss is commonly used in image reconstruction and restoration tasks~\cite{dong2015compression,svoboda2016compression,MaoSY16a}, and is
This kind of approach has shown to be effective to recover the low frequency details from a compressed image, but on the other hand most of the high frequency details are suppressed. 

\subsubsection{SSIM Loss}

The Structural Similarity (SSIM)~\cite{wang2004image} has been proposed an alternative to MSE and Peak Signal-to-Noise Ration (PSNR) image similarity measures, which have both shown to be inconsistent with the human visual perception of image similarity.
Given images $I$ and $J$, SSIM is defined as follows:\vspace{-12pt}

\begin{equation}
SSIM\left(I,J\right)=\frac{\left(2\mu_I\mu_J+C_1\right)\left(2\sigma_{IJ}+C_2\right)}{\left(\mu_I^2+\mu_J^2+C_1\right)\left(\sigma_I^2 + \sigma_J^2+C_2\right)}
\end{equation}

We optimize the training of the network with respect to the structural similarity between the uncompressed images and the reconstructed ones. Since the SSIM function is differentiable, we can define the SSIM loss as:
\begin{equation}
l_{SSIM} = -\frac{1}{WH}\sum_{x=1}^{W}\sum_{y=1}^{H}SSIM\left(I_{x,y}^{HQ}, I_{x,y}^{RQ}\right)
\end{equation}

Note that we minimize $-SSIM\left(I^{HQ}, I^{RQ}\right)$ instead of $ 1 - SSIM\left(I^{HQ}, I^{RQ}\right)$ since the gradient is equivalent.

\subsection{Generative Adversarial Artifact Removal}

The generative network architecture, defined in Sect.~\ref{sec:generative} can be used in an adversarial framework, if coupled with a discriminator. Adversarial training \cite{goodfellow2014generative} is a recent approach that has shown remarkable performances to generate synthetic photo-realistic images in super-resolution tasks \cite{Ledig-2016}. 

The aim is to encourage a generator network $G$ to produce solutions that lay on the manifold of the real data by fooling a discriminative network $ D $.  The discriminator is trained to distinguish reconstructed  patches $I^{RQ}$ from the real ones $I^{HQ}$. To condition the generative network, we feed as positive examples $I^{HQ}|I^{LQ} $ and as negative examples  $I^{RQ}|I^{LQ}$, where $\cdot|\cdot$ indicates channel-wise concatenation. For samples of size $N\times N \times C$ we discriminate samples of size $N\times N \times 2C$.

\subsubsection{Discriminative Network}

\begin{figure*}[!t]
\centering
\textbf{Discriminator Network}\vspace{10pt}\\
\includegraphics[width=0.7\textwidth]{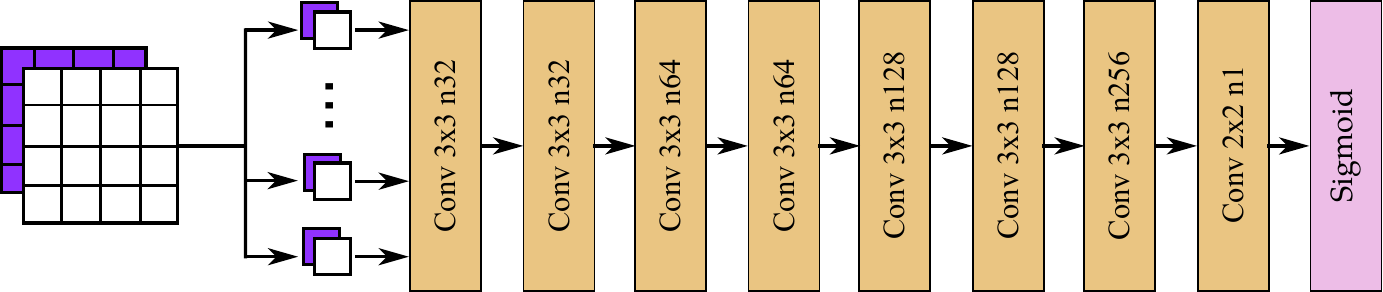}
\caption{Architecture of Discriminator Networks indicating with $n$ the number of filters for each Convolutional Layer. White squares indicate real ($I^{HQ}$) or  generated patches ($I^{RQ}$), while purple ones are their respective compressed versions $I^{LQ}$. }\vspace{-16pt}
\label{fig:Discr}
\end{figure*}

Our discriminator architecture uses  convolutions without padding with single-pixel stride and uses LeakyReLU activation after each layer. Every two layers, except the last one, we double the filters. We do not use fully connected layers. Feature map size decreases as a sole effect of convolutions reaching unitary dimension at the last layer. A sigmoid is used as activation function.  The architecture of the discriminator network is shown in Fig.\ref{fig:Discr}.

\begin{figure}[]
\centering
\includegraphics[width=.44\columnwidth]{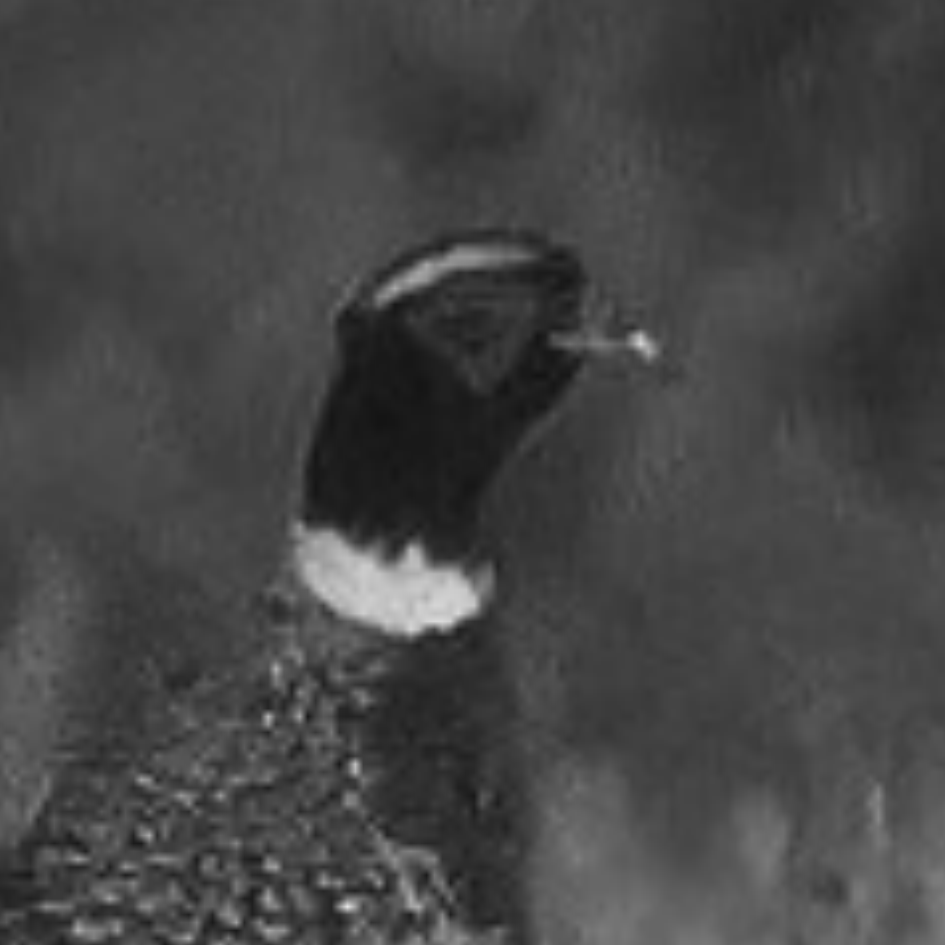}
\hspace{3pt}
\includegraphics[width=.44\columnwidth]{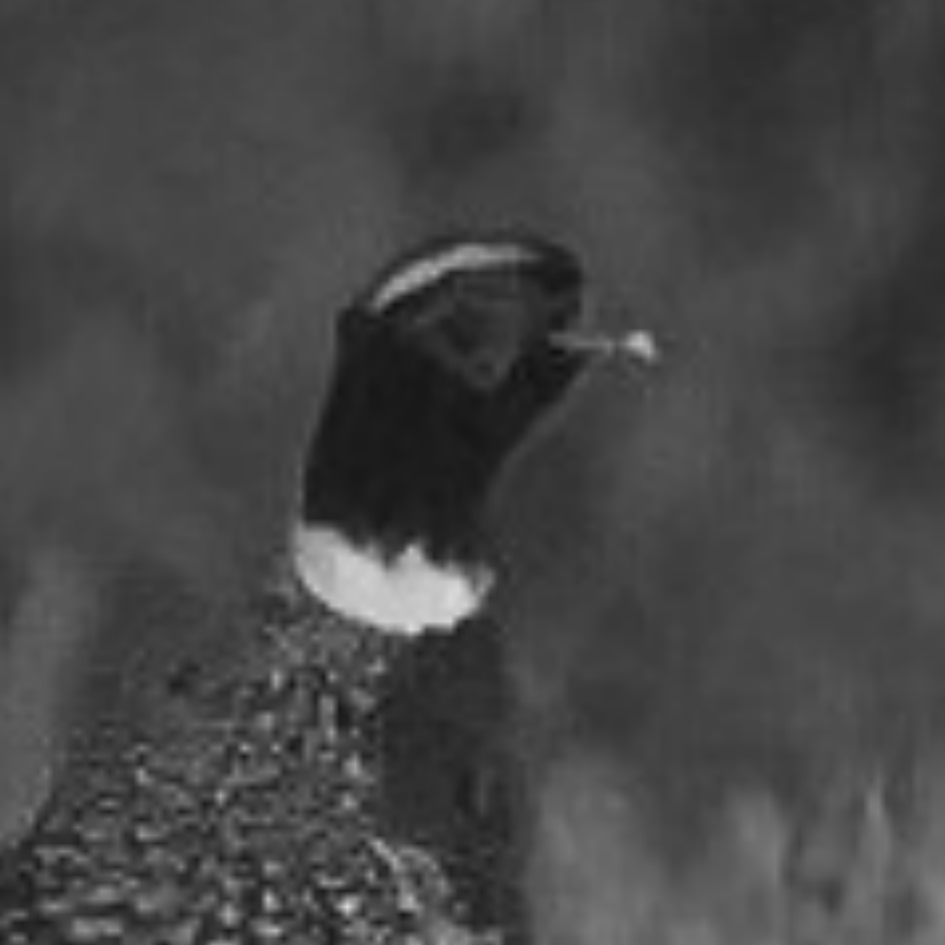}
\caption{Left: reconstruction without sub-patch strategy. Right: our sub-patch strategy reduces mosquito noise and ringing artifacts.}\vspace{-15pt}
\label{fig:GANvsPGAN}
\end{figure}

The set of weights $ \psi $ of the D network are learned by minimizing:\vspace{-12pt}
\begin{align}
l_d =& -\log\left(D_\psi\left(I^{HQ}|I^{LQ}\right)\right)\notag\\ &-\log\left(1 - D_\psi\left(I^{RQ}|I^{LQ}\right)\right)
\end{align}
Discrimination is performed at the sub-patch level, as indicated in Fig.~\ref{fig:Discr}, this is motivated by the fact that compression algorithms decompose images into patches and thus artifacts are typically created within them.
Since we want to encourage to generate images with realistic patches, $I^{HQ}$ and $I^{RQ}$ are partitioned into $P$ patches of size $ 16 \times 16$ and then they are fed into the discriminative network. In Figure~\ref{fig:GANvsPGAN} it can be seen the beneficial effect of this approach  in the reduction of mosquito noise.

\subsubsection{Perceptual Loss}
Following the contributions of Dosovitskiy and Brox \cite{DosovitskiyB16}, Johnson \textit{et al.} \cite{johnson2016perceptual}, Bruna \cite{BrunaSL15} and Gatys \cite{GatysEB15} we use a loss based on perceptual similarity in the adversarial training. The distance between the images is not computed in image space: $I^{HQ}$ and $I^{RQ}$ are initially projected on a feature space by some differentiable function $ \phi $, then the Euclidean distance is computed between the feature representation of the two images:

\begin{equation}
l_{P} =  \frac{1}{W_fH_f}\sum_{x=1}^{W_f}\sum_{y=1}^{H_f}\left(\phi\left(I^{HQ}\right)_{x,y} - \phi\left(I^{RQ}\right)_{x,y}\right)^2\label{eq:vgg}
\end{equation}
where $ W_f $ and $ H_f $ are respectively the width and the height of the feature maps.
The model optimized with the perceptual loss generates reconstructed images that are not necessarily accurate according to the pixel-wise distance measure, but on the other hand the output will be more similar from a feature representation point of view.
\subsubsection{Adversarial Patch Loss}

In the present work we used the pre-trained VGG19 model \cite{SimonyanZ14a}, extracting the feature maps obtained from the second convolution layer before the last max-pooling layer of the network.  We train the generator using the following loss:

\begin{equation}
l_{AR}  = l_{P} + \lambda l_{adv}.
\end{equation}
 Where $l_{adv}$ is the standard adversarial loss:
\begin{equation}
l_{adv} = -\log\left(D_\psi\left(I^{RQ}|I^{LQ}\right)\right)
\end{equation}
clearly rewarding solutions that are able to \dquote{fool} the discriminator.

\section{Experiments}

\subsection{Implementation Details}
All the networks have been trained with a NVIDIA Titan X GPU using random patches from MSCOCO~\cite{lin2014microsoft} training set. For each mini-batch we have sampled 16 random $ 128 \times 128 $ patches, with flipping and rotation data augmentation. We compress images with MATLAB JPEG compressor at multiple QFs, to learn a more generic model.
For the optimization process we used Adam \cite{KingmaB14} with momentum 0.9 and a learning rate of $ 10^{-4} $. The training process have been carried on for $ 70,000 $ iterations. 
In order to ensure the stability of the adversarial training we have  followed the guidelines described in \cite{SalimansGZCRC16}, performing the one-sided label smoothing for the discriminator training. 

\subsection{Dataset and Similarity Measures}
We performed experiments on two commonly used  datasets: LIVE1 \cite{liveDatabase} and the validation set of BSD500 \cite{martin2001database} using JPEG as compression.
For a fair comparison with the state-of-the art methods, we report evaluation of PSNR, PSNR-B \cite{yim2011quality} and SSIM measures for the JPEG quality factors 10, 20, 30 and 40. We further evaluate perceptual similarity through a subjective study on BSD500. Finally we use PASCAL VOC07\cite{everingham2015pascal} to benchmark object detector performance for different reconstruction algorithms.

\subsection{Comparison with State-of-the-Art}
We first evaluate the performance of our generative network trained without the adversarial approach, testing the effectiveness of our novel architecture and the benefits of SSIM loss in such training. 
For this comparison we have reported the results of our deep residual networks with skip connections trained with the baseline MSE loss and with the proposed SSIM loss. We compare our performance with the JPEG compression and three state-of-the-art approaches: SA-DCT \cite{foi2007pointwise}, AR-CNN from Dong \etal \cite{dong2015compression} and the work described by Svoboda \etal \cite{svoboda2016compression}.
In Table~\ref{Tab:QualityLIVE1} are reported the results of our approaches respectively on BSD500 and LIVE1 datasets compared to the other state-of-the-art methods for the JPEG restoration task. The results confirm that our method outperforms the other approaches for each quality measure. Specifically, we have a great improvement of PSNR and PSNR-B for the networks trained with the classic MSE loss, while as expected the SSIM measure improves a lot in every evaluation when the SSIM loss is chosen for training.

\begin{figure*}
\centering
\newcommand{\comparisonheight}{60pt}
\begin{tabular}{ccccc}
\includegraphics[height=\comparisonheight, width=.19\textwidth]{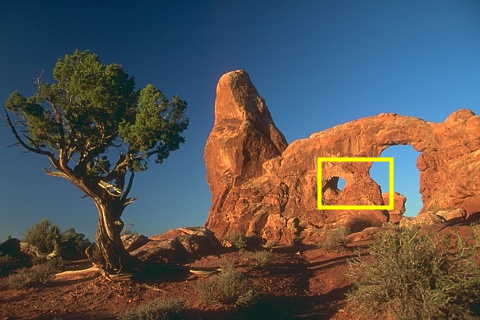}&
\includegraphics[height=\comparisonheight]{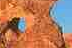}&
\includegraphics[height=\comparisonheight]{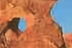}&
\includegraphics[height=\comparisonheight]{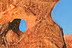}&
\includegraphics[height=\comparisonheight]{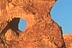}\\
\includegraphics[height=\comparisonheight,width=.19\textwidth]{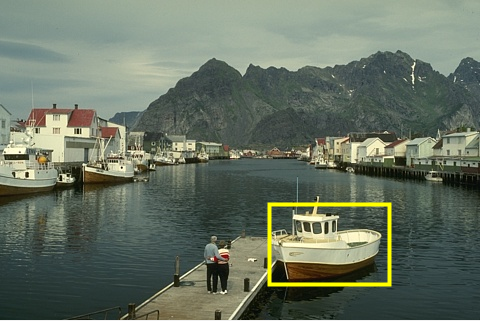}&
\includegraphics[height=\comparisonheight]{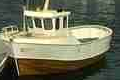}&
\includegraphics[height=\comparisonheight]{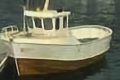}&
\includegraphics[height=\comparisonheight]{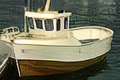}&
\includegraphics[height=\comparisonheight]{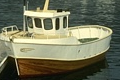}\\
Original& JPEG 20 & ARCNN & Our GAN & Original Detail\\
\end{tabular}
\caption{\label{fig:qualitative} Qualitative results shown on two complex textured details. JPEG compression introduces severe blocking, ringing and color quantization artifacts. ARCNN is able to slightly recover but produces a blurry result.  Our reconstruction is hardly discernible from the original image.}
\vspace{-12pt}

\end{figure*}

 Regarding GAN, we can state that the performance is much lower than the standard approach from a quality index point of view. However, the generated images are perceptually more convincing for human viewers as can be seen in Fig.~\ref{fig:qualitative}. 
  Further confirmation will be given in Sect.~\ref{sec:mos}, in a subjective study. The combination of perceptual and adversarial loss is responsible of generating realistic textures rather than the smooth and poor detailed patches of the MSE/SSIM based approaches. In fact, MSE and SSIM metrics tend to evaluate better more conservative blurry averages over more photo realistic details, that could be added slightly displaced with respect to their original position, as observed also in super-resolution tasks~\cite{Dahl-2017}.

\begin{table}[!htb]
\centering\vspace{-12pt}
\caption{Average PSNR, PNSR-B  and SSIM results on BDS500 and LIVE1. Evaluation using luminance.}
\label{Tab:QualityLIVE1}
\resizebox{.95\columnwidth}{!}{
\begin{tabular}{|c||l|c|c|c||c|c|c|}

\hline
\multirow{2}{*}{QF} & \multirow{2}{*}{Method}				  &\multicolumn{3}{c||}{LIVE1}& \multicolumn{3}{c|}{BSD500}\\
 &            & PSNR  & PSNR-B & SSIM  & PSNR  & PSNR-B & SSIM    \\ \hline \hline
10             & JPEG             & 27.77 & 25.33  & 0.791 & 27.58 & 24.97  & 0.769 \\
               & SA-DCT      \cite{foi2007pointwise}     & 28.65 & 28.01  & 0.809 & - & -& -\\
               & AR-CNN  \cite{dong2015compression}         & 29.13 & 28.74  & 0.823&28.74 & 28.38  & 0.796 \\
               & L4  \cite{svoboda2016compression}             & 29.08 & 28.71  & 0.824 & 28.75 & 28.29  & 0.800 \\
               & Our MSE		  & \textbf{29.45} & \textbf{29.10}  & 0.834 & \textbf{29.03} & \textbf{28.61}  & 0.807 \\
               & Our SSIM     & 28.94 & 28.46  & \textbf{0.840} & 28.52 & 27.93  & \textbf{0.816} \\ 
               & Our GAN &	27.29&	26.69&	0.773	&  27.01&	26.30&	0.746\\\hline
20             & JPEG             & 30.07 & 27.57  & 0.868  & 29.72 & 26.97  & 0.852 \\
               & SA-DCT   \cite{foi2007pointwise}        & 30.81 & 29.82  & 0.878  & - & -& -\\
               & AR-CNN    \cite{dong2015compression}        & 31.40 & 30.69  & 0.890 & 30.80 & 30.08  & 0.868\\
               & L4  \cite{svoboda2016compression}              & 31.42 & 30.83  & 0.890 & 30.90 & 30.13  & 0.871 \\
               & L8  \cite{svoboda2016compression}              & 31.51 & 30.92  & 0.891 & 30.99 & 30.19  & 0.872 \\
               & Our MSE 		  & \textbf{31.77} & \textbf{31.26}  & 0.896 & \textbf{31.20} & \textbf{30.48}  & 0.876 \\
               & Our SSIM         & 31.38 & 30.77  & \textbf{0.900} & 30.79 & 29.92  &\textbf{0.882}\\ 
               & Our GAN & 28.35&	28.10&	0.817&	  28.07	&27.76&	0.794 \\\hline
30             & JPEG             & 31.41 & 28.92  & 0.900 & 30.98 & 28.23&0.886\\
               & SA-DCT  \cite{foi2007pointwise}         & 32.08 & 30.92  & 0.908& - & -& -\\
               & AR-CNN  \cite{dong2015compression}          & 32.69 & 32.15  & 0.917& - & -&- \\
			   & Our MSE & \textbf{33.15} & \textbf{32.51}  & 0.922 & \textbf{32.44} & \textbf{31.41}&0.906\\
               & Our SSIM     & 32.87 & 32.09  & \textbf{0.925}& 32.15 & 30.97&\textbf{0.909} \\ 
               & Our GAN & 28.58&	28.75&	0.832& 28.5 & 28.00&0.811\\\hline
40             & JPEG             & 32.35 & 29.96  & 0.917 & 31.88 & 29.14&0.906\\
               & SA-DCT    \cite{foi2007pointwise}       & 32.99 & 31.79  & 0.924& - & -& -\\
               & AR-CNN   \cite{dong2015compression}         & 33.63 & 33.12  & 0.931& - & -&- \\
               & Our MSE & \textbf{34.09} & \textbf{33.40}  & 0.935& \textbf{33.30} & \textbf{32.18}& 0.921\\
               & Our SSIM     & 33.82 & 33.00  & \textbf{0.937}& 33.04 & 31.72& \textbf{0.924}\\ 
               & Our GAN 	  & 28.99 & 28.84  &	0.837& 28.61 &28.20 &0.815\\\hline
\end{tabular}\vspace{-8pt}
}
\end{table}
\subsection{Object Detection}
We are interested in understanding how a machine trained object detector performs depending on the quality of an image, in term of compression artifacts. Compressed images are degraded, and object detection performance degrades, in some cases even dramatically when strong compression is applied. In this experiment we use Faster R-CNN~\cite{ren2015faster} as detector and report results on different versions of PASCAL VOC2007; results are reported in Tab.~\ref{tab:pascal}. As an upper bound we report the mean average precision (mAP) on the original dataset. As a lower bound we report performance on images compressed using JPEG with quality factor set to 20 ($6,7\times$ less bitrate).  Then we benchmark object detection on reconstructed versions of the compressed images, comparing AR-CNN~\cite{dong2015compression}, our generative MSE and SSIM trained generators with the GAN. First of all, it must be noted that the decrease in the overall mAP measured on compressed images with respect to the upper bound is large: 14.2 points. AR-CNN,  MSE and SSIM based generators are not recovering enough information yielding around 2.1, 2.4 and 2.5 points of improvements respectively. As can be observed in Table~\ref{tab:pascal} our GAN artifact removal restores the images in a much more effective manner yielding the best result increasing the performance by 7.4 points, just 6.8 points less than the upper bound.

Our GAN artifact removal process recovers impressively on \textit{cat} (+16.6), \textit{cow} (+12.5), \textit{dog} (+18.6) and \textit{sheep} (+14.3), which are classes where the object is highly articulated and texture is the most informative cue. In these classes it can also be seen that MSE and SSIM generators are even deteriorating the performance, as a further confirmation that the absence of higher frequency components alters the recognition capability of an object detector.
To assess the effect of color we report the use of GAN using only luminance (GAN-Y). Using $l_P$ defined as in Eq.~\ref{eq:vgg} is important, switching to a simpler L1 loss (GAN-L1) we obtain much lower performance.  Our GAN trained with a full patch discriminator obtains .605 mAP, while our sub-patch discriminator leads to .623 mAP, highlighting its importance.

\begin{figure}
\centering
\includegraphics[width=.8\columnwidth]{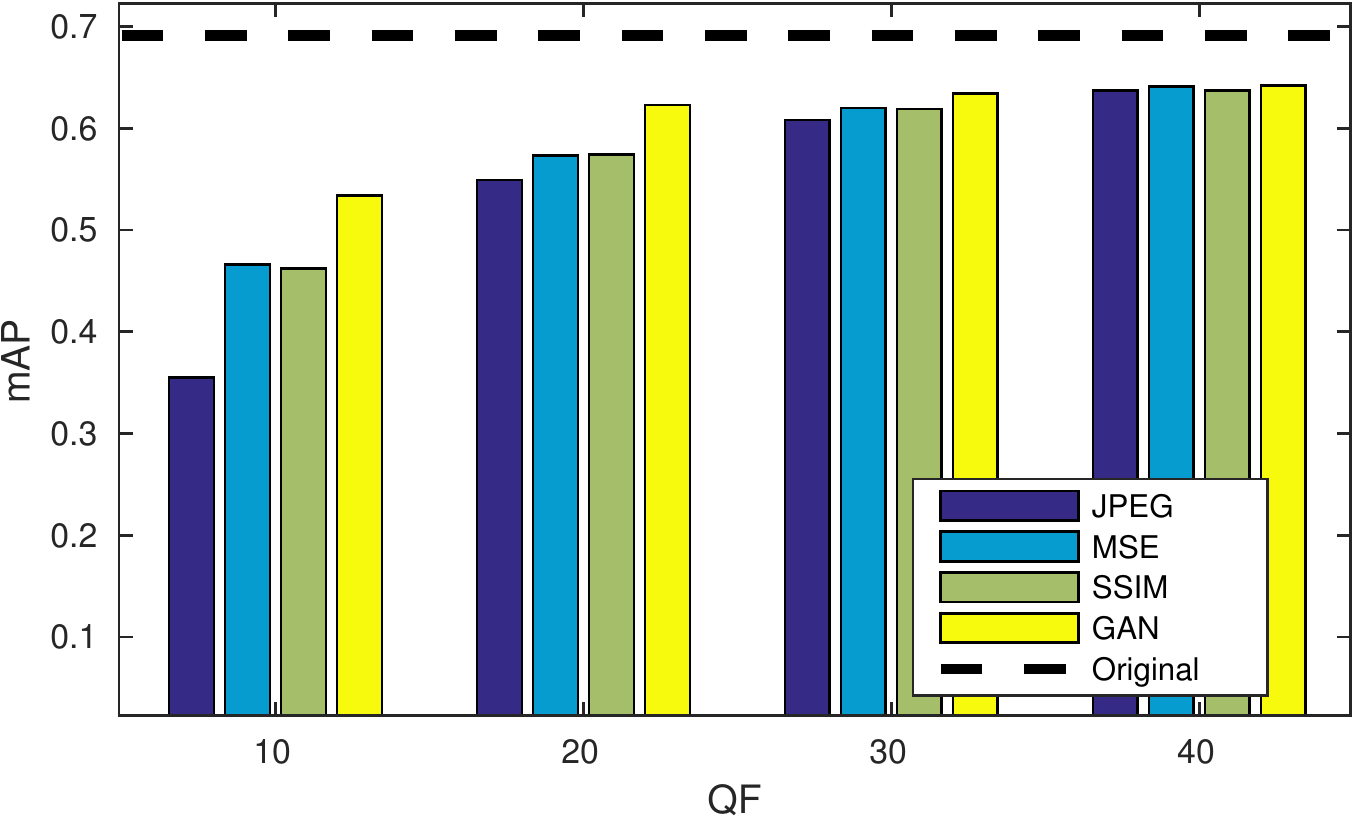}
\caption{Mean average precision (mAP), for different Quality Factors (QF), and restoration approaches, on PASCAL VOC2007.\label{fig:mapqf}}\vspace{-24pt}
\end{figure}

In Fig.~\ref{fig:mapqf} we analyze the effects of different compression levels, changing the quality factor.
GAN is able to recover details even for very aggressive compression rates, such as QF=10. In Fig.~\ref{fig:mapqf} it can be seen how GAN always outperform other restoration algorithms. The gap in performance is reduced when QF raises, e.g QF=40 ($4,3\times$ less bitrate). 

\newcommand{\icon}[1]{\includegraphics[height=12pt]{./images/icons/result/#1}}
\begin{table*}[!htb]
\centering
\resizebox{.75\textwidth}{!}{
\begin{tabular}{lccccccccccc}
&\icon{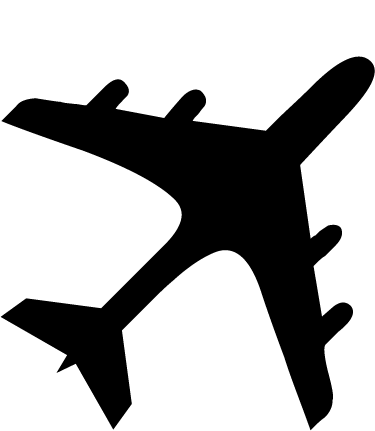} & \icon{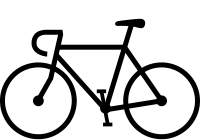} & \icon{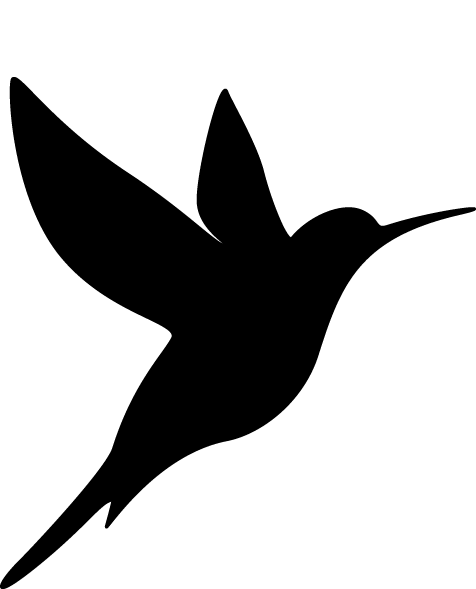}      & \icon{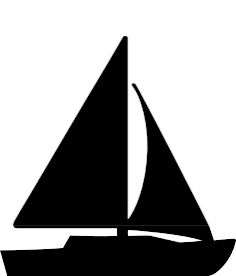}   & \icon{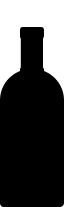} & \icon{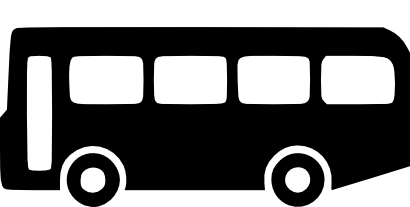}    & \icon{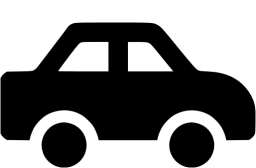}    & \icon{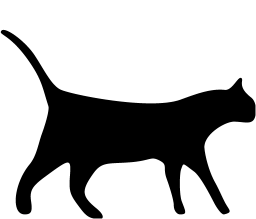}    & \icon{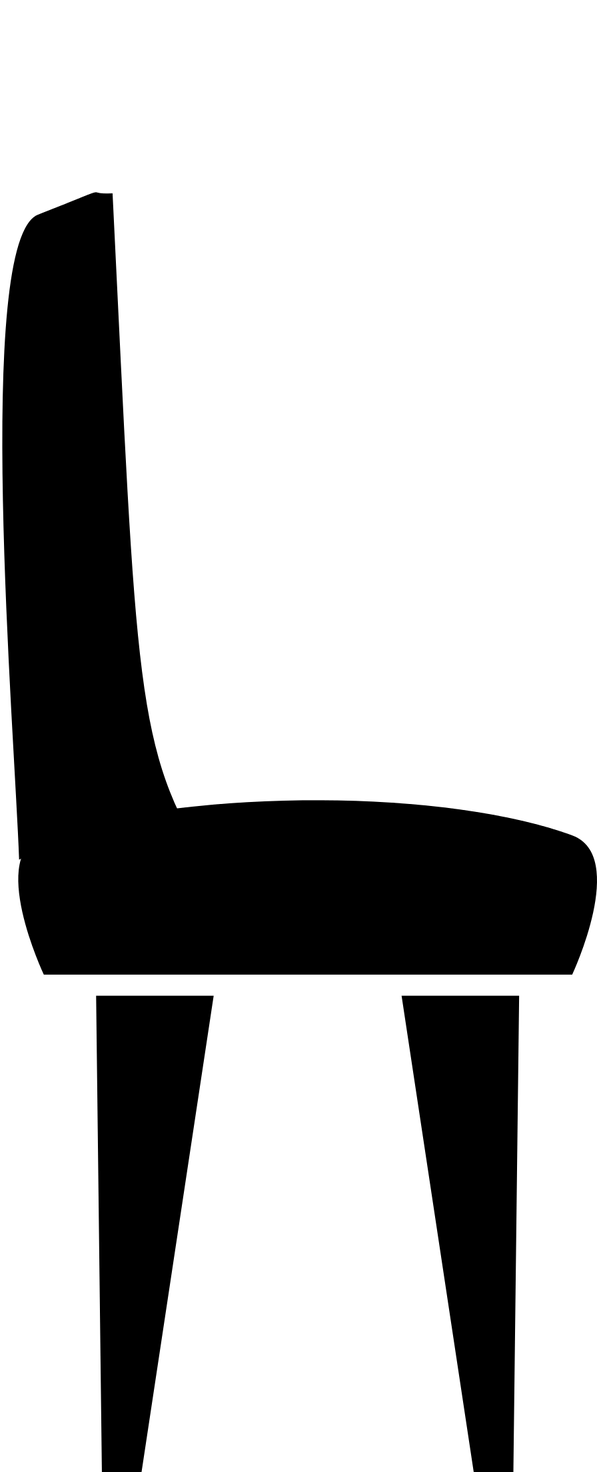}  & \icon{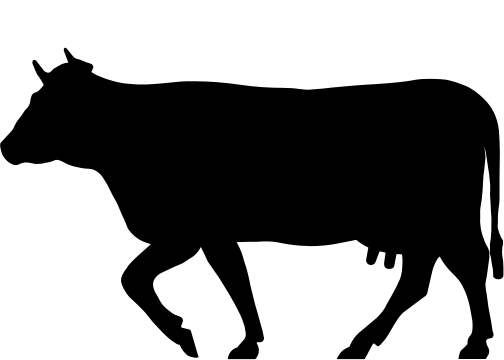} & \icon{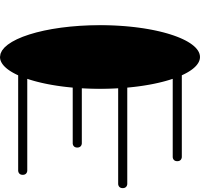}     \\ \hline
JPEG 20         & 0.587   & 0.692 & 0.516 & 0.434  & 0.350 & 0.673 & 0.71  & 0.559 & 0.334 & 0.559       & 0.579 \\\hline
AR-CNN \cite{dong2015compression} & 0.641 &	0.686&	0.523&	0.413&	0.367&	0.702&	0.742&	0.530&	0.363&	0.574&	0.607\\\hline	MSE          & 0.647   & 0.696 & 0.512 & 0.406  & 0.409 & 0.713 & 0.75  & 0.542 & 0.386 & 0.546       & 0.614 \\\hline
Our SSIM         & 0.655   & 0.706 & 0.513 & 0.417  &\textbf{ 0.411} & 0.713 & 0.746 & 0.555 & 0.387 & 0.538       & \textbf{0.615} \\\hline
Our GAN-Y & 0.657   & 0.696 & 0.547 & 0.461  & 0.354 & 0.719 & 0.708 & 0.673 & 0.380  & 0.653       & 0.605 \\\hline
Our GAN-L1 & 0.644	& 0.75	& 0.524	& 0.421	& 0.427	& 0.691	& 0.755	& 0.667	& 0.402	& 0.616	& 0.597 \\\hline
Our GAN      & \textbf{0.666}   & \textbf{0.753} & \textbf{0.565} &\textbf{ 0.475}  & {0.395} &\textbf{ 0.727} & \textbf{0.770}  & \textbf{0.725} & \textbf{0.403} & \textbf{0.684}       & {0.602} \\\hline
Original            & 0.698   & 0.788 & 0.692 & 0.559  & 0.488 & 0.769 & 0.798 & 0.858 & 0.487 & 0.762       & 0.637\\\hline
& \icon{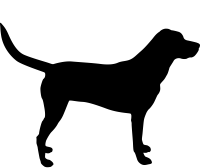} &  \icon{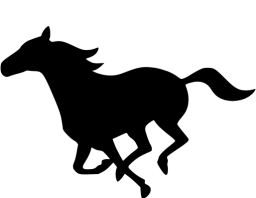} &  \icon{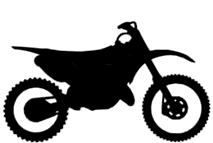} &  \icon{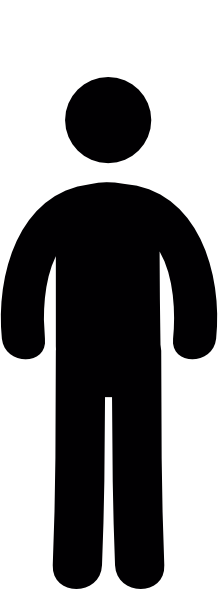} &  \icon{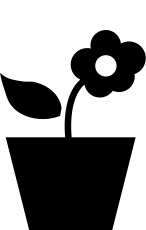} &  \icon{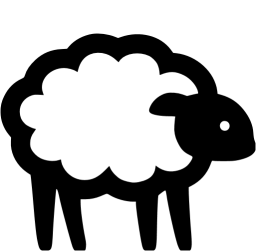} & \icon{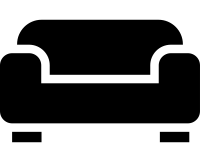} &  \icon{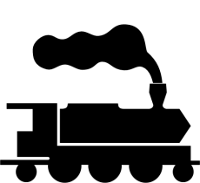} &  \icon{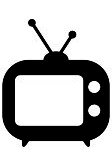} & \multicolumn{2}{c}{\textbf{mAP}} \\ \hline
JPEG 20         & 0.532 & 0.691 & 0.665 & 0.638 & 0.260  & 0.482 & 0.434 & 0.707 & 0.570  & \multicolumn{2}{c}{0.549} \\\hline
AR-CNN \cite{dong2015compression} & 0.581&	0.724	&0.661&	0.658&	0.313&	0.499&	0.526&	0.712&	0.578&	\multicolumn{2}{c}{0.570}\\\hline

Our MSE          & 0.595 & 0.713 & 0.668 & 0.664 & \textbf{0.310}  & 0.485 & 0.522 & 0.676 & 0.600   & \multicolumn{2}{c}{0.573} \\\hline
Our SSIM         & 0.596 & 0.720  & 0.666 & 0.663 & 0.308 & 0.482 & 0.532 & 0.668 & 0.598 & \multicolumn{2}{c}{0.574} \\\hline
Our GAN-Y & 0.681 & 0.738 & 0.661 & 0.662 & 0.290  & 0.608 & 0.544 & 0.722 & 0.600   & \multicolumn{2}{c}{0.598} \\\hline
Our GAN-L1 & 0.679	& 0.749	& 0.666	& 0.664	& 0.309	& 0.543	& 0.587	& 0.655	& 0.613	& \multicolumn{2}{c}{0.598}\\\hline
Our GAN     & \textbf{0.718} & \textbf{0.753} & \textbf{0.707} & \textbf{0.670}  & 0.303 & \textbf{0.625} & \textbf{0.586} & \textbf{0.712} & \textbf{0.611} & \multicolumn{2}{c}{\textbf{0.623}} \\\hline
Original            & 0.790  & 0.802 & 0.757 & 0.763 & 0.376 & 0.683 & 0.672 & 0.777 & 0.667 & \multicolumn{2}{c}{0.691}
\end{tabular}
}
\caption{Object detection performance measured as mean average precision (mAP) on PASCAL VOC2007 for different reconstruction algorithms. Bold numbers indicate best results among reconstruction approaches.\label{tab:pascal} }\vspace{-12pt}
\end{table*}

To assess the generality of our approach we tested it on other codecs: WebP, BPG and JPEG2000. We tuned all codecs to obtain the same average bitrate on the whole VOC2007 dataset of the respective JPEG codec using a QF of 20. The improvement in mAP is using our GAN is the following WebP(+4\%), JPEG2000 (+2.9\%) and BPG (+2.1\%).


\subsection{Subjective evaluation}\label{sec:mos}

\begin{figure}[]
\centering
\includegraphics[width=.8\columnwidth]{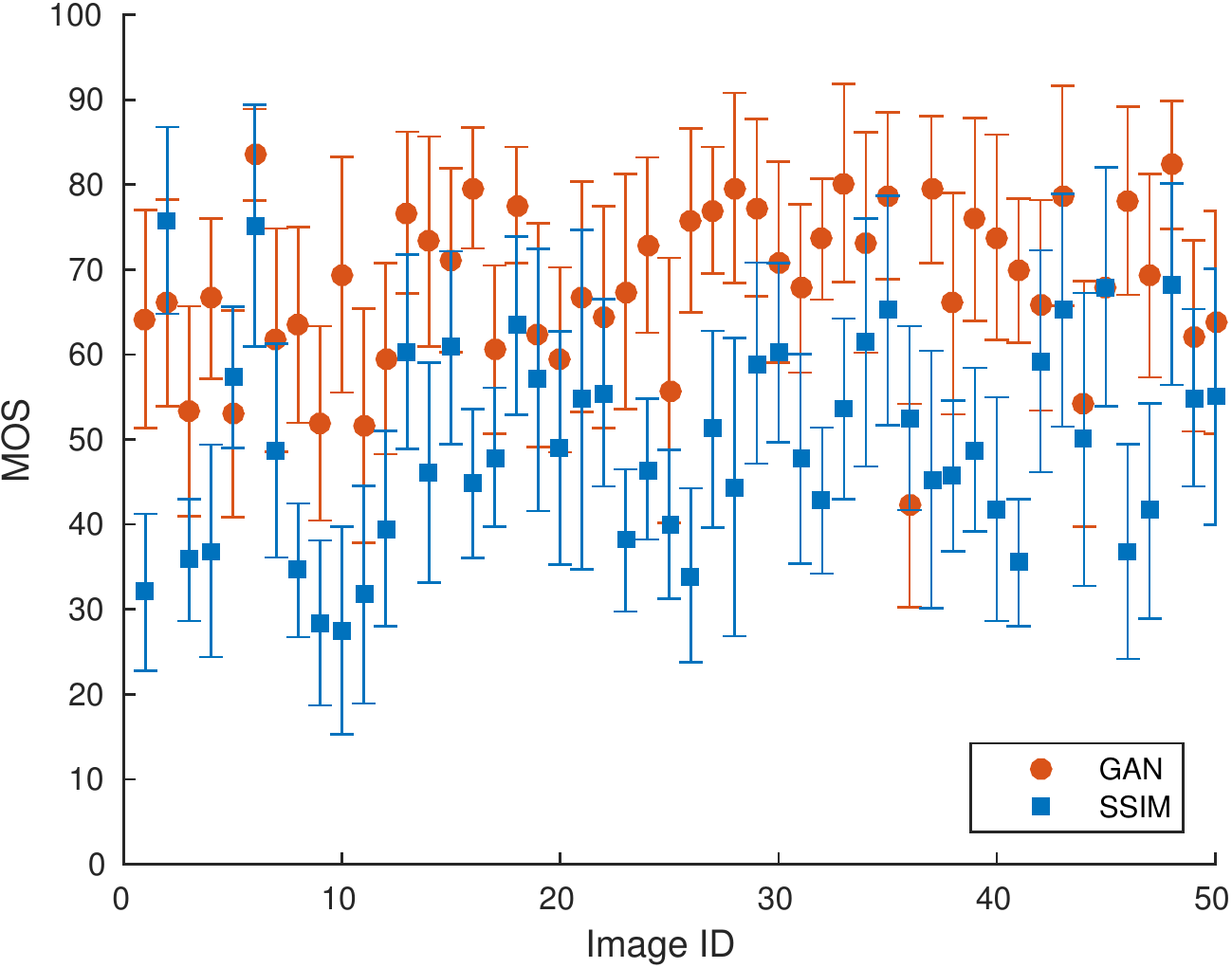}\par\smallskip
\caption{MOS values, with 0.95 confidence, for all the 50 images used in the subjective evaluation. }
\label{fig:subjective_evaluation}
\vspace{-18pt}
\end{figure}

In this experiment we evaluate how images processed with the proposed methods are perceived by a viewer, comparing in particular how the SSIM loss and the GAN-based approaches preserve the details and quality of an image. We have recruited 10 viewers, a number that is considered enough for subjective image quality evaluation tests \cite{Winkler-2009}; none of the viewers was familiar with image quality evaluation or the work presented in this paper. Evaluation has been done following a DSIS (Double-Stimulus Impairment Scale) setup, created using VQone, a tool specifically designed for this type of experiments \cite{VQone-2016}: subjects evaluated the test image in comparison to the original image, and graded how similar is the test image to the original, using a continuous scale from 0 to 100, with no marked values to avoid choosing preferred numbers. We have randomly selected 50 images from the BSD500 dataset, containing different subjects, such as nature scenes, man-made objects, persons, animals, etc. For each original image both an image processed with the SSIM loss network and the GAN network have been shown, randomizing their order to avoid always showing one of the two approaches in the same order, and randomizing also the order of presentation of the tests for each viewer. The number of 50 images has been selected to maintain the duration of each evaluation below half an hour, as suggested by ITU-R BT.500-13 recommendations \cite{ITU-R-2012} (typical duration was $\sim20$ minutes). Overall 1,000 judgments have been collected and final results are reported in Table \ref{Tab:subjective_comparison} as MOS (Mean Opinion Scores) with standard deviation. Results show that the GAN-based network is able to produce images that are perceived as more similar to the original image. A more detailed analysis of results is shown in Fig.~\ref{fig:subjective_evaluation}, where for each image is reported its MOS with $95\%$ confidence. It can be observed that in $90\%$ of the cases the images restored with the GAN-based network are considered better than using the SSIM-based loss. Fig.~\ref{fig:subjective_evaluation} shows two examples, one where GAN performs better (see the texture on the elephant skin) and one of the few where SSIM performs better (see the faces).

\begin{table}[]
\centering\vspace{-12pt}
\caption{Subjective image quality evaluation in terms of Mean Opinion Score(MOS)  on BSD500.}
\label{Tab:subjective_comparison}
\begin{tabular}{|l|c|c|}\hline
Method 		& MOS	&  std. dev.  \\ \hline \hline
Our SSIM	& 49.51		 	& 22.72 \\
Our GAN		& \textbf{68.32}	& 20.75 \\ \hline
\end{tabular}
\end{table}

\begin{figure}[!hbt]
\centering
\includegraphics[width=\columnwidth]{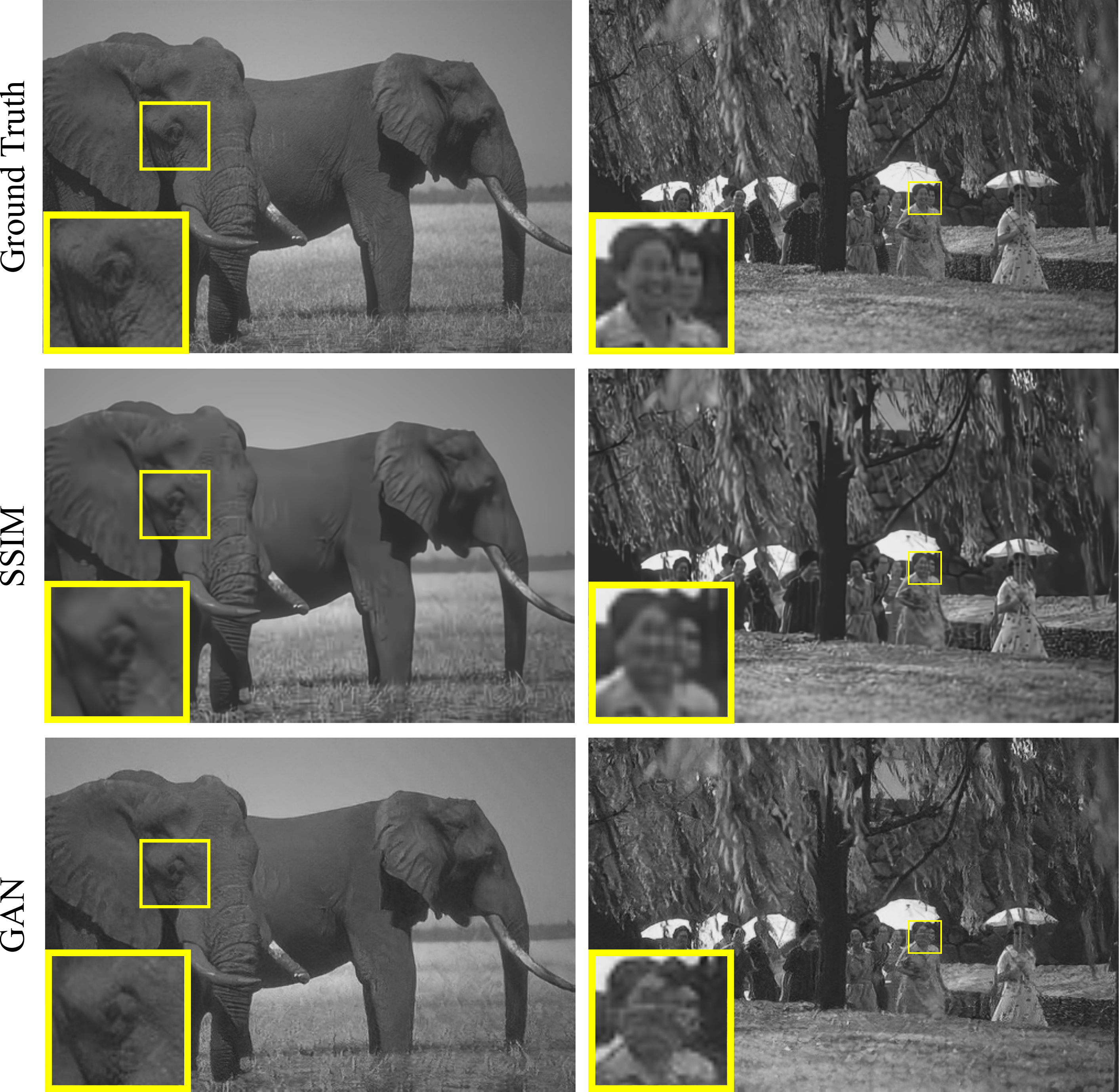} 
\caption{ Samples of BSD500 validation set used in our subjective evaluation. Left column: best result for the GAN approach, right column: best result for the SSIM approach.}\vspace{-18pt}
\end{figure}

\section{Conclusion}

We have shown that it is possible to remove compression artifacts by transforming images with deep convolutional residual networks. Our generative network trained using SSIM loss obtains state of the art results according to standard image similarity metrics. Nonetheless, images reconstructed as such appear blurry and missing details at higher frequencies. These details make images look less similar to the original ones for human viewers and harder to understand for object detectors. We therefore propose a conditional Generative Adversarial framework which we train alternating full size patch generation with sub-patch discrimination. Human evaluation and quantitative experiments in object detection show that our GAN generates images with finer consistent details and these details make a difference both for machines and humans.

\section*{Acknowledgments}
We gratefully acknowledge the support of NVIDIA Corporation with the donation of the Titan X Pascal GPU used for this research.

\bibliographystyle{ieee}
\bibliography{egbib}
\balance

\end{document}